\documentclass[preprint,12pt]{elsarticle}

\usepackage{amsmath,amsfonts,amssymb}
\usepackage{graphicx}
\usepackage{caption}
\usepackage{subcaption}

\newtheorem{thm}{Theorem}

\newtheorem{rem}{Remark}
\newproof{pf}{Proof}
\newproof{pot}{Proof of Theorem \ref{Th1}}

\renewcommand{\P}{\mathbb P}

\newcommand{\C}{\mathbb C}
\newcommand{\D}{\mathbb D}
\newcommand{\R}{\mathbb R}
\renewcommand{\H}{\mathbb H}
\newcommand{\eps}{\epsilon}

\newcommand{\SE}{\mathrm{SE3}}
\newcommand{\ci}{\mathrm{i}}
\newcommand{\qi}{\mathbf{i}}
\newcommand{\qj}{\mathbf{j}}
\newcommand{\qk}{\mathbf{k}}

\journal{Mechanism and Machine Theory}

\begin{document}

\begin{frontmatter}

\title{A Technique for Deriving Equational Conditions on the 
	Denavit-Hartenberg Parameters of 6R Linkages that are Necessary
	for Movability}

\author[zlimainaddress]{Zijia Li\corref{zlicorrespondingauthor}}
\cortext[zlicorrespondingauthor]{Corresponding author. Tel: +43 (0)732 2468 5253 \ Fax: +43 (0)732 2468 5212 }
\ead{zijia.li@oeaw.ac.at}

\author[jschichomainaddress]{Josef Schicho}
\ead{josef.schicho@risc.jku.at}

\address[zlimainaddress]{Johann Radon Institute for
  Computational and Applied Mathematics Austrian Academy of Sciences
  (RICAM), Altenbergerstrasse 69, 4040 Linz, Austria}

\address[jschichomainaddress]{Research Institute for Symbolic Computation Johannes Kepler University, 
  Altenberger Straße 69, A-4040 Linz, Austria}

\begin{abstract}

A closed 6R linkage is generically rigid. Special cases may be mobile. Many families 
 of mobile 6R linkages have been characterised in terms of the invariant Denavit-Hartenberg 
 parameters of the linkage. In other words, many {\em sufficient conditions} for mobility are known. 
 In this paper we give, for the first time, equational 
 conditions on the invariant Denavit-Hartenberg parameters that are {\em necessary} for mobility.
 The method is based on the theory of bonds. We illustrate the method by deriving the equational
 conditions for various well-known linkages (Bricard's line symmetric linkage, Hooke's linkage,
 Dietmaier's linkage, and recent a generalization of Bricard's orthogonal linkage),
 starting from their bond diagrams; and by deriving the equations for another
 bond diagram, thereby discovering a new mobile 6R linkage.

\end{abstract}

\begin{keyword}
Denavit-Hartenberg parameters, overconstrained 6R linkages, necessary conditions, bond diagrams
\end{keyword}

\end{frontmatter}

\section{Introduction}\label{in:0}
\label{intr}

A closed $n$R (n revolute joints) linkage is uniquely determined by its set of Denavit Hartenberg parameters~\cite{Denavit1955},
which consists of $3n$ real numbers: the twist angles, the normal distances and the offsets.
If $n=6$, then a generic choice of these parameters leads
to a rigid linkage. In the literature, there are many families of special choices of parameters
such that the linkage is mobile, or in other words, several sufficient conditions for mobility
are known in \cite{B93, Dietmaier95}. 
 A recent overview on known families can be found in \cite{chenr12}.
However, we are still far away from a complete classification of all mobile 6R linkages.
In this paper, we try to make a step towards such a complete classification, by deriving necessary conditions 
for mobility; up to our knowledge, not a single necessary equation condition has been known up to now. 

Our method is based on the bond theory, which has been introduced in \cite{Hegedues13b} in order
to study overconstrained linkages. This theory provides a {\em classification scheme} for 6R linkages:
for any mobile 6R linkage, one can calculate a certain combinatoric structure describing algebraic
relations between the joints, called the {\em bond diagram}. This diagram consists of {\em bonds},
which are connections between two joints of the linkage. Any joint is connected to at least one
other joint, and adjacent joints are never connected. 
If we number the joints cyclically, then a bond connects either joints $i$ and $i+2$ --
then we speak of a {\em near connection}, -- or it connects joints $i$ and $i+3$, and then we speak of
a {\em far connection}. 

In \cite{Hegedues13b}, it was shown that the existence of a near connection implies the validity
of a well-known condition which also arises frequently in many families, namely Bennett's condition:
$s_i=0$, and $\frac{d_i}{\sin(\phi_{i})}=\frac{d_{i+1}}{\sin(\phi_{i+1})}$, where the
$s_i,d_i,\phi_i$ are Denavit-Hartenberg parameters (see section~\ref{pr:0} for the precise definitions).
Bennett's condition is equivalent to a kinematic condition on three consecutive rotation axes,
not all three parallel or intersecting,
namely the existence of a fourth axis such that the closed 4R linkage with these four axes is movable
(see \cite{bennett14}).
However, there are many mobile 6R linkages without near connections, for instance Bricard's orthogonal
linkage or Bricard's line symmetric linkage. So, the Bennett conditions are not necessary
for mobility.

The paper \cite{Hegedues13b} contains no equational condition implied by the existence for a far connection.
The main contribution of this paper fills this gap, by introducing the {\em quad polynomials}:
these are univariate polynomials of degree~2
with coefficients depending on the Denavit-Hartenberg parameters by an explicit formula.
The existence of a far connection implies a common root of two 
such quad polynomials, and this gives rise to necessary equational conditions.

Because every mobile linkage has either near or far connections (or both), it is then possible
to write down equational conditions for movability (see Remark~\ref{rem:cond}). However,
the full system of equations is too big and complicated, and therefore it is better to
follow the classification scheme suggested by bond theory and distinguish cases according
to the bond diagram. For any bond diagram,
we will derive a non-trivial system of algebraic equations, consisting of Bennett conditions
for the near connections and quad polynomial conditions for the far connections. In some cases,
the equations are even sufficient for movability, hence the equations characterize all linkages with this
particular bond diagram. 

In Subsection~\ref{sec:known}, we illustrate the method by deriving the equational conditions 
for various well-known linkages. The bond diagram studied in Subsection~\ref{ss:new} leads to
a new movable 6R linkage $L$: we show that for every known family, there is an algebraic condition which is
not satisfied by the set of Denavit-Hartenberg parameters of $L$. In order to show that $L$ is indeed
movable, we calculate the configuration set by solving the corresponding algebraic system of equations
and observe that it is one-dimensional (there is no geometric proof for mobility for this example).
We find this new family of linkages especially remarkable because two of its R-joints
can be replaced by H-joints (helical joints), and the linkage remains movable. 

\paragraph*{Original contribution of this paper}
The theory of bonds is not new: it has first been introduced in \cite{Hegedues13b}. The contribution of
this paper is comparatively modest: we just give equational conditions for the existence of far
connections. But this is not trivial, and without these equational conditions it would not be possible
to derive necessary equational conditions for movability of 6R linkages.

The linkage in Subsection~\ref{ss:new} is also original, but it is merely a side result: our main motivation
is not to invent new families of linkages but to make progress in the complete classification
of mobile 6R linkages. 

It should also be pointed out the scope of bond theory is much larger than the technique of
quad polynomials. While bond theory is applicable for a large class if linkages (e.g. multiply closed,
linkages with different types of joints), quad polynomials can only be defined for simply closed
linkages with 6 joints/links.

\paragraph*{Structure of the paper} 
The remaining part of the paper is set up as follows. In section 2, we introduce all preliminary  definitions we need. In section 3, 
we give the definition of the \emph{quad polynomial} and its main property.
Section 4 contains examples (old and new).

\section{Preliminary Definitions}\label{pr:0}
\subsection{Computation of the Configuration Space}
\label{sec:notations}

In this section we recall a method of computing the configuration space of a closed 6R linkage using
dual quaternions and Denavit-Hartenberg parameters. 
First, we start by introducing the set of Denavit-Hartenberg parameters of a closed 6R linkage.
For $i=1,\dots,6$, let $l_i$ be the rotation axis of the $i$-th joint. 
The angle $\phi_i$ is defined as the angle of the direction vectors of
$l_i$ and $l_{i+1}$ (with some choice of orientation). We also set $c_i:=\cos(\phi_i)$
and $w_i=\cot(\frac{\phi_i}{2})=\frac{cos(\phi_i)+1}{sin(\phi_i)}$.The number $d_i$ is defined as the orthogonal distance of the lines $l_i$ and $l_{i+1}$.
Note that $d_i$ may be negative; this depends on some choice of orientation of the common normal,
which we denote by $n_i$.

From now on, we will always assume that there are no parallel adjacent lines, which means
that the angles $\phi_1,\dots,\phi_6$ are not equal to $0$ or $\pi$.
Then we may set $b_i:=\frac{d_i}{\sin(\phi_i)}$ (Bennett ratios \cite{mavroidis95} (inverse)) as an abbreviation.
 Finally, we define the offset $s_i$ as the signed distance of the intersections of the common normals
$n_{i-1}$ and $n_i$ with $l_i$. 

The Denavit-Hartenberg parameters $\phi_i,d_i,s_i$ are invariant when the linkage is moving.
Moreover, it is well-known that they form a complete system of invariants for all closed 6R linkages 
without adjacent parallel lines: if two such linkages share all parameters, then there is a 
configuration that moves the first into the second. 
(A description of invariant parameters for 6-bar linkages with adjacent parallel lines 
may be found in \cite{Baker03}). 

The closure condition is an equation in the group $\SE$ of Euclidean displacements.
We give a formulation in the language of dual quaternions, based on the fact that $\SE$ is
isomorphic to the multiplicative group of dual quaternions with nonzero real norm modulo multiplication
by nonzero real scalars. The set $\D\H$ is the 8-dimensional real vector-space generated by
$1,\qi,\qj,\qk,\eps,\eps\qi,\eps\qj,\eps\qk$ with multiplication inherited from quaternion multiplication
and the rule $\eps^2=0$ (the dual number $\eps$ is considered as a scalar).
The set of dual quaternions with real norm is also called the Study quadric
and denoted by $S$; it is a quadric hypersurface in $\R^8$.

In the isomorphism described in \cite[Section~2.4]{husty10}, the rotation with axis determined by $\qi$ 
and angle $\phi$ corresponds to the dual quaternion $(\cos(\frac{\phi}{2})-\sin(\frac{\phi}{2})\qi)$,
which is projectively equivalent to $(\cot(\frac{\phi}{2})-\qi)$. The translation
parallel to $\qi$ by a distance $d$ corresponds to the dual quaternion $\left(1-\frac{d}{2}\eps\qi\right)$.
So the closure equation is
\begin{equation}\label{eq:1}
  (t_1-\qi)g_1(t_2-\qi)g_2\cdots(t_6-\qi)g_6 \in \R^\ast,
\end{equation}
where
\begin{equation}\label{gi:1}
g_i=\left(1-\frac{s_i}{2}\eps\qi\right)\left(w_i-\qk\right)\left(1-\frac{d_i}{2}\eps\qk\right),
\end{equation}
for $i=1,\dots,6$.
This is just the reformulation of the well-known closure equations \cite{Denavit1955} in terms of dual quaternions.
\begin{rem}
  In \cite{Hegedues13b,Hegedues13f,Li13a}, we used a different formulation of the closure equation, namely
\begin{equation}
  \label{eq:0}
  (t_1-h_1)(t_2-h_2)\cdots(t_6-h_6) \in \R^\ast,
\end{equation}
where $h_1,\dots,h_6$ are dual quaternions specifying the rotation axes in some initial position
of the linkage.
\end{rem}

 The set $K$ of all $6$-tuples $(t_1,\dots,t_6)$ fulfilling 
\eqref{eq:1} is called the \emph{configuration set} of the linkage $L$. 
 The dimension of the \emph{configuration set} is called the \emph{mobility} of the linkage. 
 We are mostly interested in linkages of mobility one.

For a fixed set of Denavit-Hartenberg parameters, the configuration set can be computed with
the help of the computer algebra Maple. For example, let $L$ be the Bricard line symmetric linkage with parameters
\[ (d_1,\dots,d_6) = \left(\frac{3}{5},\frac{24}{13},\frac{72}{25},
                           \frac{3}{5},\frac{24}{13},\frac{72}{25}\right), \] 
\[ (w_1,\dots,w_6) = \left(\frac{1}{3}, \frac{2}{3}, \frac{3}{4}, 
                           \frac{1}{3}, \frac{2}{3}, \frac{3}{4}\right), \]
\[ (s_1,\dots,s_6) = (4, 5, 1, 4, 5, 1) . \]
 Then all $g_i$s of this 6R linkage are
\[      g_1 =\frac{1}{3}-\frac{3}{10}\eps-\frac{2}{3}\eps \qi-2 \eps\qj-\left(\frac{1}{10}\eps+1\right)\qk,\]
\[      g_2 =\frac{2}{3}-\frac{12}{13}\eps-\frac{5}{3}\eps\qi-\frac{5}{2}\eps\qj-\left(\frac{8}{13}\eps+1\right)\qk,\]
\[      g_3 =\frac{3}{4}-\frac{36}{25}\eps-\frac{3}{8}\eps\qi-\frac{1}{2}\eps\qj-\left(\frac{27}{25}\eps+1\right)\qk,\]
\[ g_4  =g_1,\ g_5=g_2, \ g_6=g_3.\]

We expand the left hand side of the closure equation~\eqref{eq:1}. The coordinates 2,\dots,8 have to be zero,
this gives 7 polynomial equations in $t_1,\dots,t_6$. One of these equations is redundant, namely the 5th
(the coefficient of $\eps$). The reason is that the left hand side is always in the Study quadric $S$,
and if an element of the form $a+b\eps$ is in $S$ then it follows that $b=0$. In order to exclude
unwanted solutions, we add
the inequality $(t_1^2+1)(t_2^2+1)(t_3^2+1)(t_4^2+1)(t_5^2+1)(t_6^2+1)\ne 0$. In order to solve
this system with the computer program Maple, we introduce an extra variable $u$ and add the equation
$(t_1^2+1)(t_2^2+1)(t_3^2+1)(t_4^2+1)(t_5^2+1)(t_6^2+1)u-1=0$, and compute a Gr\"obner basis that
eliminates $u$ again. The solution set has some zero-dimensional components, which are not interesting,
and a one-dimensional component:
\begin{equation*}
 171 t_1^2 t_2^2-134 t_1^2 t_2+40 t_1 t_2^2+49 t_1^2-160 t_1 t_2-5 t_2^2-24 t_1+90 t_2-255,
\end{equation*}
\begin{equation*}
 171 t_1 t_2^2+19 t_2^2 t_3-134 t_1 t_2+40 t_2^2-222 t_2 t_3+49 t_1-288 t_2+105 t_3,
\end{equation*}
\begin{equation*}
 171 t_1 t_2-133 t_1 t_3+19 t_2 t_3-134 t_1+40 t_2-222 t_3-323, t_1-t_4, t_2-t_5, t_3-t_6.
\end{equation*}
\subsection{The Bond Diagram}

In this section we recall the fundamentals of bond theory introduced by \cite{Hegedues13b}. Its purpose
is to associate to every mobile linkage a diagram, which describes algebraic relations between the joints.
From this diagram one can draw conclusions, like the validity of Bennett conditions in various cases.

 Let $L$ be a closed $6$R linkage with mobility~1.
 Let $K_\C\subset(\P^1_\C)^6$ be the Zariski closure of $K$ the configuration set, that is, the
zero set of all polynomial equations that vanish on $K$, including complex points and points
at infinity. The set of bonds is defined as
\begin{equation}
  \label{eq:6}
  B := \{(t_1,\ldots,t_6) \in K_\C \mid
  (t_1-h_1)g_1(t_2-h_2)g_2\cdots(t_6-h_6)g_6 = 0\}.
\end{equation}

Let $\beta$ be a bond with coordinates $(t_1,\ldots,t_6)$. By Theorem~2 in \cite{Hegedues13b},
 there exist indices $1\leq i<j \leq 6$, such that $t_i^2 + 1 = t_j^2 + 1 = 0$. 
 If there are exactly two coordinates of $\beta$ with values $\pm \ci$ (where $\ci$ denotes the  
 imaginary unit in the field of complex numbers $\mathbb{C}$), then
 we say that $\beta$ {\em connects} joints $i$ and $j$. By \cite[Corollary\ 12]{Hegedues13b}, we have
\begin{equation}\label{bond:eqs}
 (t_i-h_i)g_i(t_{i+1}-h_{i+1})g_{i+1}\cdots(t_{j}-h_j)=0 .
\end{equation}
In general, the situation is more complicated: a bond may connect several pairs of joints, or
it may connect a single pair of joints with higher connection multiplicity;
we refer to \cite{Hegedues13b} for the technical details in these cases.

We visualize bonds and their connection numbers by \emph{bond diagrams.}
 We start with the link diagram, where vertices correspond to links and edges
 correspond to joints. Then we draw connections between edges for the bonds connecting
the corresponding joints. Note that bonds always come in pairs of complex conjugates,
and conjugate bonds connect the same joints; so we draw only one line for each conjugate pair of bonds.

\begin{figure*}[!htb]
        \centering
        \begin{subfigure}[b]{0.250\textwidth}
                \centering
                \includegraphics[width=\textwidth]{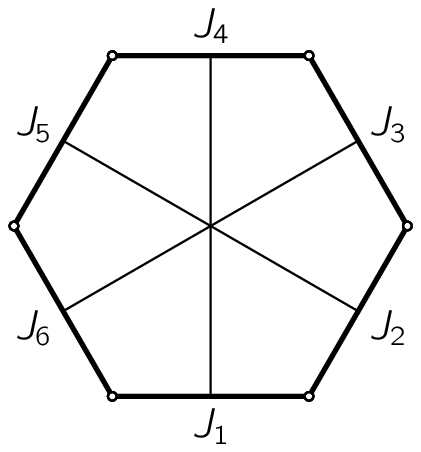}
                \caption{}
                \label{fig:1}
        \end{subfigure}\hspace{1cm}
        \begin{subfigure}[b]{0.250\textwidth}
                \centering
                \includegraphics[width=\textwidth]{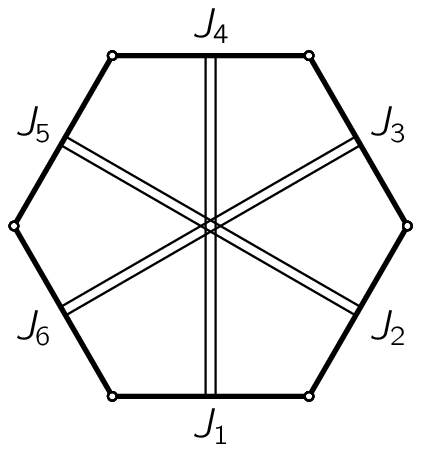}
                \caption{}
                \label{fig:2}
        \end{subfigure}\hspace{1cm}
	\begin{subfigure}[b]{0.250\textwidth}
                \centering
                \includegraphics[width=\textwidth]{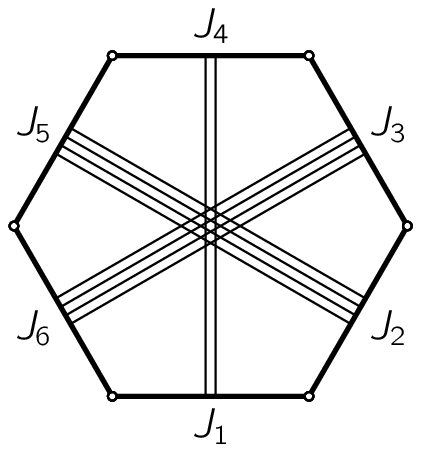}
                \caption{}
                \label{fig:3}
        \end{subfigure}\\
        \begin{subfigure}[b]{0.250\textwidth}
                \centering
                \includegraphics[width=\textwidth]{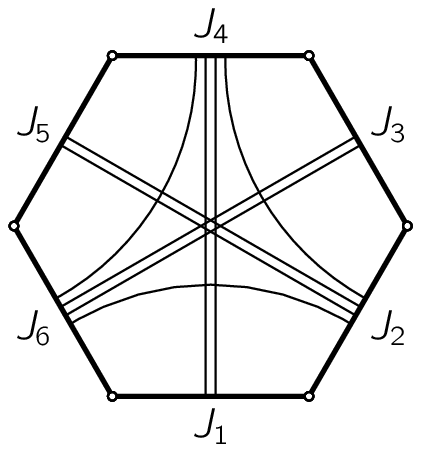}
                \caption{}
                \label{fig:4}
        \end{subfigure}\hspace{1cm}
        \begin{subfigure}[b]{0.250\textwidth}
                \centering
                \includegraphics[width=\textwidth]{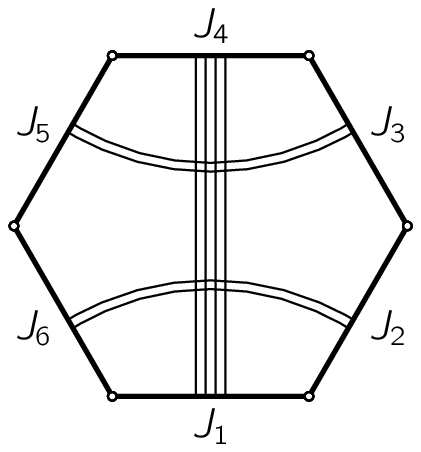}
                \caption{}
                \label{fig:5}
        \end{subfigure}\hspace{1cm}        
        \begin{subfigure}[b]{0.250\textwidth}
                \centering
                \includegraphics[width=\textwidth]{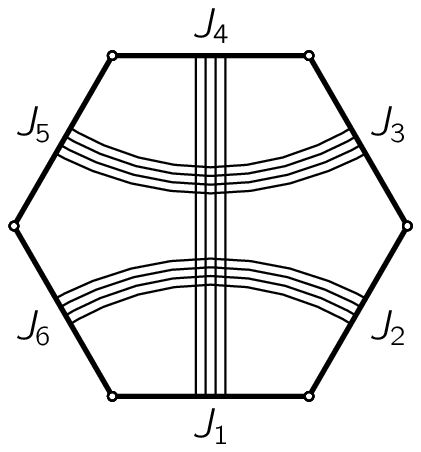}
                \caption{}
                \label{fig:7}
        \end{subfigure}\\
        \begin{subfigure}[b]{0.250\textwidth}
                \centering
                \includegraphics[width=\textwidth]{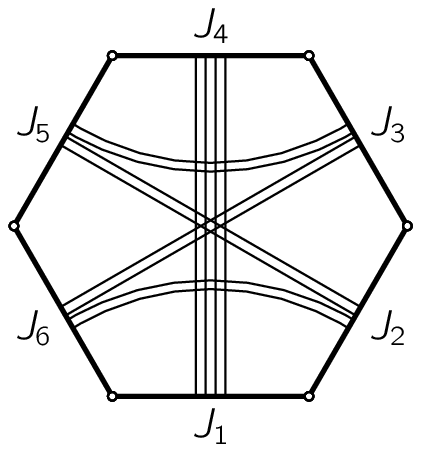}
                \caption{}
                \label{fig:8}
        \end{subfigure}\hspace{1cm}
        \begin{subfigure}[b]{0.250\textwidth}
                \centering
                \includegraphics[width=\textwidth]{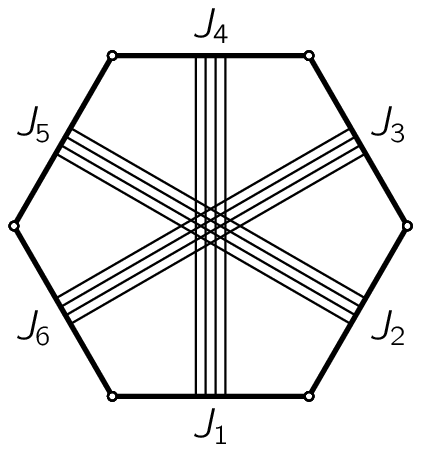}
                \caption{}
                \label{fig:9}
        \end{subfigure}
        \caption{Bond diagrams for 
    the Cube linkage (a), the Bricard line symmetric linkage (b), the new linkage (c),
            the Waldron partially symmetric linkage (d), the Bricard plane symmetric linkage (e), 
        the Hooke linkage (f), the Dietmaier linkage (g), 
        and the Orthogonal Bricard linkage (h). The joints are
	labeled by $J_1,\dots,J_6$. Each bond connects two joints.
	}\label{fig:bonds}
\end{figure*}

In order to draw the bond diagram for a given linkage, we first compute its configuration space
as in the previous section. If $K$ is a Gr\"obner bases for the configuration space, then we calculate
for any pair $(i,j)$ of indices a Gr\"obner bases $K_{ij}$ of the union of $K$ and $\{t_i^2+1,t_j^2+1\}$.
The number of bonds connecting joints $J_i,J_j$ is then the degree of the ideal $K_{ij}$.
In Figure~\ref{fig:bonds}, we show some known examples with bond diagrams.

We number the joints cyclically by $J_1,\dots,J_6$.
By \cite[Theorem 3(c)]{Hegedues13b}, a bond cannot connect $J_i$ and $J_{i+1}$ 
(with addition modulo 6 because of cyclic indexing). We speak of a {\em near connection}
if a bond connects joints $J_i$ and $J_{i+2}$, and of a {\em far connection} if a bond connects
$J_{i}$ and $J_{i+3}$. For instance, the bond diagram in Figure~\ref{fig:bonds}(d)
has 3 near connections and 6 far connections.

\begin{thm} \label{thm:bennett}
A near connection implies a Bennett condition: if $J_i$ and $J_{i+2}$ are connected, then
$b_i=\pm b_{i+1}$ and $s_{i+1}=0$.
\end{thm}

\begin{pf}
This is an immediate consequence of Theorem~1 and Corollary~2 of \cite{Hegedues13b}.
\end{pf}

For any two links, their relative motion can be described by a curve in the Study quadric.
The degree of this curve can be read off the bond diagram, using \cite[Theorem 5]{Hegedues13b}.
We will not use this theorem in its full generality. The only consequence which we will use
is that every joint is connected to at least one other joint. Otherwise the relative motion of
the two adjacent links would have to be of degree $0$, which means a constant, which means
that this joint would have to be frozen throughout the motion of the linkage.

\begin{rem} \label{rem:isomorphism_bond}
There might exist different 6R linkages which have same bond diagram. For instance, 
 there are two different families of 6R linkages corresponding to the bond 
 diagram Figure~\ref{fig:bonds}(h) \cite{Hegedues15g}. The 6R linkage with translation property in 
 \cite{Li13p} has the same bond diagram as the Bricard line symmetric linkage. 
 An interesting diagram is Figure~\ref{fig:bonds}(e). It is the bond diagram of the Bricard 
 plane symmetric 
 linkage. 
 But there may be another family of 6R linkage with the same bond diagram,
 we do not
  know.

\end{rem}

\section{Quad Polynomials}\label{quad:0}

In this section, we introduce a technique to derive algebraic equations on the Denavit-Hartenberg parameters 
which are necessary for the existence of a far connection. Because any mobile linkage has
either a near or a far connection, this allows to deduce necessary conditions for movability.

Assume that $\beta=(\ci,\alpha,\beta,\ci,\alpha',\beta')$ is a bond connecting $J_1$ and $J_4$
(note that the first and fourth coordinate must be $\pm\ci$ by the definition of a connection).
We define $q:=(\ci-\qi)g_1(\alpha-\qi)g_2(\beta-\qi)g_3$. Then we have
\[ (\ci+\qi)q = 0 = q(\ci-\qi), \]
where the first equation follows from $\ci^2-\qi^2=0$ and the second equation is a consequence
of bond theory. All solutions of the two equations above are scalar multiples of $\qj+\ci\qk$,
the scalar being an arbitrary dual number. Hence we may write $q=(a+b\eps)(\qj+\ci\qk)$ with
unique numbers $a,b\in\C$. If $a\ne 0$ (this is the case when both $\alpha$ and $\beta$ are not equal to $\pm \ci$), 
 then $q$ is projectively equivalent (that is, up to multiplication by a nonzero complex number) 
 to $(1+x_0\eps)(\qj+\ci\qk)$ for some $x_0\in\C$. Our goal is to
define a polynomial $Q_1^+\in\C[x]$, with coefficients depending on the Denavit-Hartenberg parameters,
such that $Q_1^+(x_0)=0$.

The triple of complex numbers $(\alpha,\beta,x_0)$ satisfies the following three equations:
\begin{enumerate}
\item the first coordinate (the coefficient of $1$) of $q$ is zero;
\item the fifth coordinate (the coefficient of $\eps$) of $q$ is zero;
\item the product of the third coordinate (the coefficient of $\qj$) of $q$ and $x$
	is equal to the seventh coordinate (the coefficient of $\eps\qj$) of $q$.
\end{enumerate}
Conversely, if a triple $(\alpha,\beta,x_0)$ satisfies these three equations, it is
straightforward to show that $q$ as defined above is projectively equivalent to $(1+x_0\eps)(\qj+\ci\qk)$.
We define now the quad polynomial $Q_1^+(x)$ as the resultant of the three polynomial equations
above with respect to the variables $\alpha,\beta$. In order to achieve uniqueness, we
assume that $Q_1^+$ is normed, that is, its leading coefficient is~1. 
Using the computer algebra system Maple, the computation of the resultant can be done
with symbolic expressions for the Denavit-Hartenberg parameters. The result is the quadric polynomial
\[ Q_1^+(x) = \left(x+\frac{b_{3}c_3-b_{1}c_1}{2}-\frac{s_{1}}{2}\ci\right)^2 + \]
 \[ \frac{\ci}{2}\left(b_1 s_{2}+b_{3} s_{3}+s_{2} b_{3} c_{2}+s_{3} b_1 c_{2}\right) -\]
 \[ \frac{b_1 b_{3} c_{2}-s_{2} s_{3} c_{2}}{2}
   + \frac{s_{2}^2+s_{3}^2-b_1^2+b_{2}^2-b_{3}^2-b_{2}^2 c_{2}^2}{4}. \]
For $i=2,\dots,6$, we define the quad polynomial $Q_i^+(x)$ by a cyclic shift of indices that
shifts $1$ to $i$. Finally, we define $Q_i^-(x)$ by replacing the parameters $c_1,\dots,c_6,b_1,\dots,b_6$
and $s_2,s_4,s_6$ by their negatives, and leaving $s_1,s_3,s_5$ as they are. For instance,
\[ Q_1^-(x) = \left(x+\frac{b_{3}c_3-b_{1}c_1}{2}-\frac{s_{1}}{2}\ci\right)^2 + \]
 \[ \frac{\ci}{2}\left(b_1 s_{2}-b_{3} s_{3}-s_{2} b_{3} c_{2}+s_{3} b_1 c_{2}\right) -\]
 \[ \frac{-b_1 b_{3} c_{2}-s_{2} s_{3} c_{2}}{2}
   + \frac{s_{2}^2+s_{3}^2-b_1^2+b_{2}^2-b_{3}^2-b_{2}^2 c_{2}^2}{4}. \]

\begin{thm} \label{thm:q}
Let $k$ be the number of bond connections of $J_1$ and $J_4$. Then
\[ k \le \deg(\gcd(Q_1^+,Q_4^+)) + \deg(\gcd(Q_1^-,Q_4^-)) . \]
\end{thm}

\begin{pf}
Assume that $\beta=(\pm\ci,\alpha,\pm\beta,\pm\ci,\alpha',\beta')$ is a bond connecting $J_1$ and $J_4$.
We may assume that its first coordinate is $\ci$;
otherwise, we replace $\beta$ by its complex conjugate.
As above, we define $q:=(\ci-\qi)g_1(\alpha-\qi)g_2(\beta-\qi)g_3$. By the construction of the quad polynomial $Q_1^+$, there exists $x_0\in\C$ such that
$q$ is projectively equivalent to $(1+x_0\eps)(\qj+\ci\qk)$ and $Q_1^+(x_0)=0$.
Now we apply a cyclic shift and obtain, in the same way, an $x_1\in\C$ such that
$q':=(\ci-\qi)g_4(\alpha'-\qi)g_5(\beta'-\qi)g_6$ is projectively equivalent to $(1+x_1\eps)(\qj+\ci\qk)$
and $Q_4^+(x_1)=0$. Now $\beta$ satisfies all algebraic equations that are valid in the
configuration set, in particular the equation expressing that 
$(t_1-\qi)g_1(t_2-\qi)g_2(t_3-\qi)g_4$ is projectively equivalent to the quaternion conjugate of
$(t_4-\qi)g_4(t_5-\qi)g_5(t_6-\qi)g_6$. Hence $q$ and $q'$ are conjugate as dual quaternions,
up to complex scalar multiplication. But the scalar parts of both $q$ and $q'$ vanish,
hence $q$ and $q'$ are projectively equivalent. Hence $x_0=x_1$, and we have derived
the existence of a common zero of $Q_1^+$ and $Q_4^+$, under the assumption of the existence
of a bond with $t_1=t_4=\ci$. Hence $\deg(\gcd(Q_1^+,Q_4^+))$ is an upper bound for the
number of bond connections of $J_1$ and $J_4$ by bonds with $t_1=t_4=\ci$. Similarly, one shows that 
$\deg(\gcd(Q_1^-,Q_4^-))$ is an upper bound for the
number of bond connections of $J_1$ and $J_4$ by bonds with $t_1=-t_4=\ci$. 
\end{pf}

\begin{rem} \label{rem:cond}
It is well-known that two univariate polynomials have a greatest common divisor of positive degree 
if and only their resultant is zero. The resultant of two quad polynomials is a polynomial expression
in the Denavit-Hartenberg parameters. Its vanishing gives rise to two equations, because the resultant
has a real and an imaginary part. If the product of all these resultants times, say, the product
of all offsets is not zero, then no bond can exist and the linkage is rigid.
\end{rem}

The polynomial conditions obtained in the way described above are big and difficult to solve. It is
therefore more promising to go through some case distinctions on the bond diagram. In order to
obtain the strongest possible algebraic conditions, one should make the following assumptions
on the bond structure.

\begin{itemize}
\item For each of the six pairs $(J_i,J_{i+2})$ of near joints, we make an assumption whether they 
	are connected or not.
\item For each pair of the three pairs $(J_i,J_{i+3})$ of far joints, we make an assumption
	on the number of connections by bonds with $t_i=t_{i+3}=\ci$ and on the number of connections
	by bonds with $t_i=-t_{i+3}=\ci$. In both cases, this number is in the set $\{0,1,2\}$.
\item The assumptions must be consistent with the condition that every joint is attached to at least
	one bond. This condition is a consequence of \cite[Theorem 5]{Hegedues13b}, assuming that every
	joint is moving.
\end{itemize}

In the case when the number of connections of $(J_i,J_{i+3})$ by bonds with $t_i=t_{i+3}=\ci$
is equal to 2, then the two polynomials $Q_i^+(x)$ and $Q_{i+3}^+(x)$ must be equal, because
 they are both quadratic and normed and have a quadratic greatest common divisor. This is equivalent
to the vanishing of four polynomials in the Denavit-Hartenberg parameters, namely the real
and the complex part of the linear and the constant coefficient of the difference polynomial.

\section{Examples}\label{ex:0}

In the first subsection, we apply the method of quad polynomials to several well-known families
of mobile 6R linkages. The main purpose of this subsection is to show that our method
is another way to ``explain'' already known equations with a unifying method. 
In the second subsection, we use the method to obtain a new family of mobile 6R linkages.
This family is remarkable because two of its R-joints may be replaced by H-joints, and
the linkage is still movable.

As in the section~\ref{pr:0}, we use the Bennett ratios $b_1,\dots,b_6$, the angle cosines $c_1,\dots,c_6$
and the offsets $s_1,\dots,s_6$. In addition, we also use the values $f_k=c_kb_k$, $k=1,\dots,6$,
as abbreviations; this leads to shorter formulas.

\subsection{Some Known Examples}\label{sec:known}

\subsubsection{Bricard's Line Symmetric Linkage}

If $b_i=b_{i+3}$, $w_i=w_{i+3}$, and $s_i=s_{i+3}$ for $i=1,2,3$, then there is a one-dimensional
set of line symmetric positions which allow the link to move. Apparently, we also have
\[ Q_1^+=Q_4^+, \ Q_2^+=Q_5^+, \ Q_3^+=Q_6^+ , \]
which means that the necessary conditions for a double connection between each pair of far joints
are satisfied. As we saw in Figure~\ref{fig:bonds}(b), the bond diagram does indeed have these
three double connections.

Conversely, it is not true that the 12 equations (obtained by $Q_{i}^+=Q_{i+3}^+$ for $i=1,2,3$) imply that the linkage is
line symmetric. A counterexample is Bricard's orthogonal linkage, see section~\ref{ss:bo} below.

\subsubsection{Hooke's Double Spherical Linkage}

By combining two spherical linkages with one joint in common, and then removing the common joint,
we obtain a movable 6R linkage that has two triples of three joint axes meeting in a point 
(say the axes of $J_6,J_1,J_2$ and the axes of $J_3,J_4,J_5$). In this case, it is easy to see that
$b_1=b_3=b_4=b_6=s_1=s_4=0$. Another equation, namely
\[ s_2^2+s_3^2+b_2^2-f_2^2+2 s_2s_3c_2=s_5^2+s_6^2+b_5^2-f_5^2+2 s_5s_6c_5 \]
can be derived by geometric considerations (see \cite{Dietmaier95}) or 
an algebraic method (see \cite{Cui11}). 
 Alternatively, we consider
the bond diagram of Hooke's linkage, which is shown in Figure~\ref{fig:bonds}(f). As we have a fourfold
connection of $J_1$ and $J_4$, we get $Q_1^+=Q_4^+$ and $Q_1^-=Q_4^-$, and under the assumption
that $b_1=b_3=b_4=b_6=s_1=s_4=0$, this is equivalent to the above condition.

\subsubsection{Dietmaier's Linkage}

In \cite{Dietmaier95}, Dietmaier describes a family of mobile 6R linkages, which he found  by a computer-supported
numerical seach. It can be characterised by the equations
\[ b_6=b_1,b_3=b_4,b_2=b_5,c_2=c_5,f_6+f_1=f_3+f_4, \] 
\[s_6=s_2,s_3=s_5,s_1=s_4=0 . \]
Its bond diagram is shown in Figure~\ref{fig:bonds}(g).

Starting from the assumption on the bond structure, we first obtain
the conditions $b_6=b_1$, $b_3=b_4$, $s_1=s_4=0$ as consequences of Bennett conditions
implied by the existence of short connections. 
Since we have again a fourfold connection of
$J_1$ and $J_4$, we again get $Q_1^+=Q_4^+$ and $Q_1^-=Q_4^-$. We added the inequality
condition $b_1b_4\ne 0$ and did a computer-supported analysis of the solution set using Maple. 
It turns out that there are two components. The first one is Dietmaier's family.
The second is given by the equations
\[ b_6=b_1,b_3=b_4,b_2=-b_5,c_2=c_5,f_6+f_1=f_3+f_4,\]
\[s_6=s_2,s_3=s_5,s_1=s_4=0 . \]
We computed the configuration set of a random instance of the second component. It appeared
to be finite, hence the second component is not a family of mobile 6R linkages. However,
the subset of solutions that also fulfill the condition $f_1+f_6=0$ is a well-known
family, namely the Bricard plane symmetric linkage. Its bond diagram is shown in Figure~\ref{fig:bonds}(e).

\subsubsection{Bricard's Orthogonal Linkage} \label{ss:bo}

The well-known family (see \cite{Baker80}) of orthogonal linkages can be described by the conditions
\[ s_1=\dots=s_6=0, \ c_1=\dots=c_6=0, \]
\[ b_1^2+b_3^2+b_5^2=b_2^2+b_4^2+b_6^2 . \]
(The name of this family already tells the twist angles are right angles.) It is easy to
prove that these equations imply 
\begin{equation} \label{eq:444}
  Q_1^+=Q_4^+, Q_2^+=Q_5^+, Q_3^+=Q_6^+, Q_1^-=Q_4^-, Q_2^-=Q_5^-, Q_3^-=Q_6^- ,
\end{equation}
which means that the the necessary conditions for the existence of the maximal number
of far connections is fulfilled. Indeed, the bond diagram has all these far connections; this
can be seen in Figure~\ref{fig:bonds}(h).

The system of equations~\eqref{eq:444} has more solutions, leading to other linkages
with the same bond diagram. This is studied in \cite{Hegedues15g}.

\subsection{A New Example}\label{ss:new}

We assume that there are no near connections, four connections of $J_2$ and $J_5$, four connections
of $J_3$ and $J_6$, and two connections of $J_1$ and $J_4$ by bonds with $t_1=t_4$. The bond
diagram can be seen in Figure~\ref{fig:bonds}(c). 
 Then we get the following equalities of polynomials in $\C[x]$:
\begin{equation} \label{eq:442}
  Q_1^+=Q_4^+, Q_2^+=Q_5^+, Q_3^+=Q_6^+, Q_3^-=Q_6^-, Q_2^-=Q_5^-. 
\end{equation}

Using the computer algebra system Maple, we obtained the following equivalent system of solutions:
$$ b_1^2+b_3^2+b_5^2+f_6^2=b_2^2+b_4^2+b_6^2+f_3^2, f_2+f_3=f_5+f_6, $$
$$ b_2 c_1-b_3=b_2 c_3-b_1=b_5 c_4-b_6=b_5 c_6-b_4=0, $$
$$ s_2= s_3= s_5= s_6=0, s_1=s_4. $$

The solution set is irreducible. Here is a random numerical example:
$$ b_1=-\frac{1}{3}, b_2=-\frac{61}{33}, b_3=\frac{305}{429}, b_4=\frac{2000}{1001}, b_5=-\frac{2900}{1001},
   b_6=\frac{1740}{1001}, $$
$$ w_1=\frac{2}{3}, w_2=-4, w_3=\frac{6}{5}, w_4=\frac{1}{2}, w_5=\frac{\sqrt{54083849}}{6619},
   w_6=\frac{3}{7}, $$
 $$ s_2= s_3= s_5= s_6=0, s_1=s_4=\frac{2}{3}. $$

\begin{figure*}[!htb]
        \centering
        \begin{subfigure}[b]{0.250\textwidth}
                \centering
                \includegraphics[width=\textwidth]{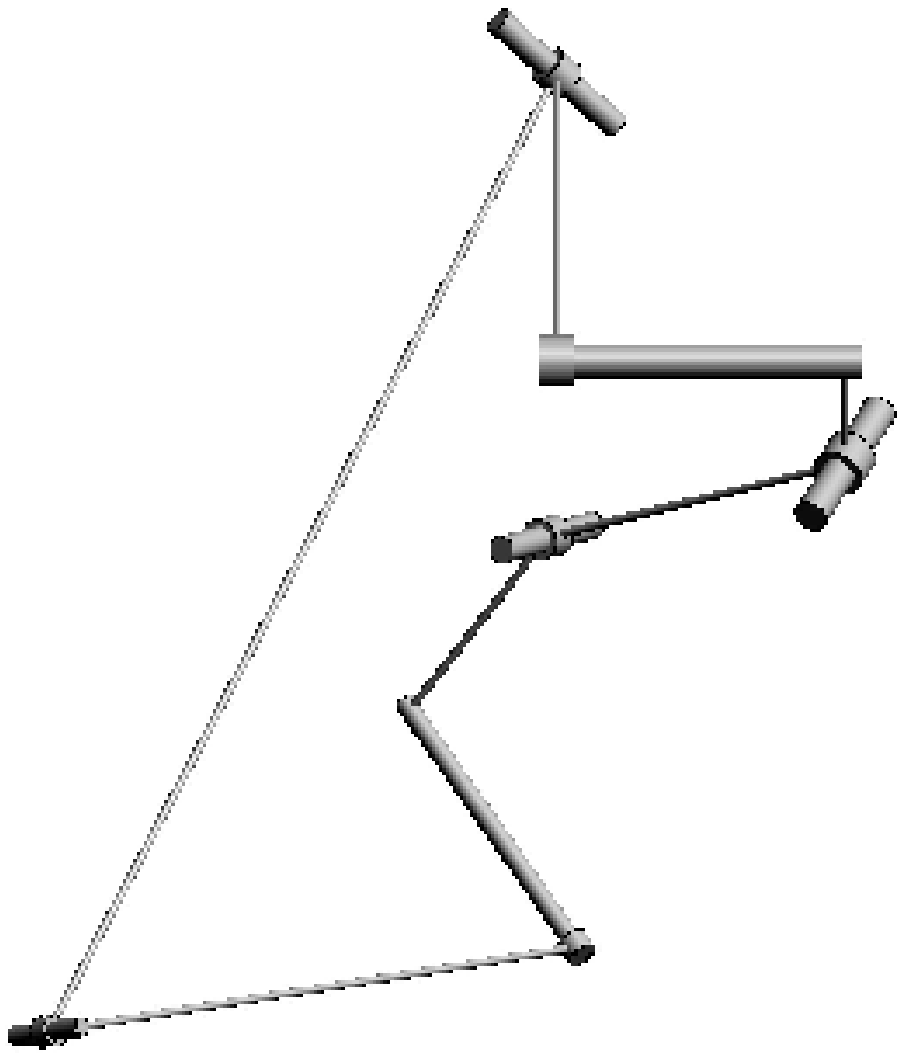}
                \caption{}
                \label{fig:a}
        \end{subfigure}\hspace{1cm}
	\begin{subfigure}[b]{0.250\textwidth}
                \centering
                \includegraphics[width=\textwidth]{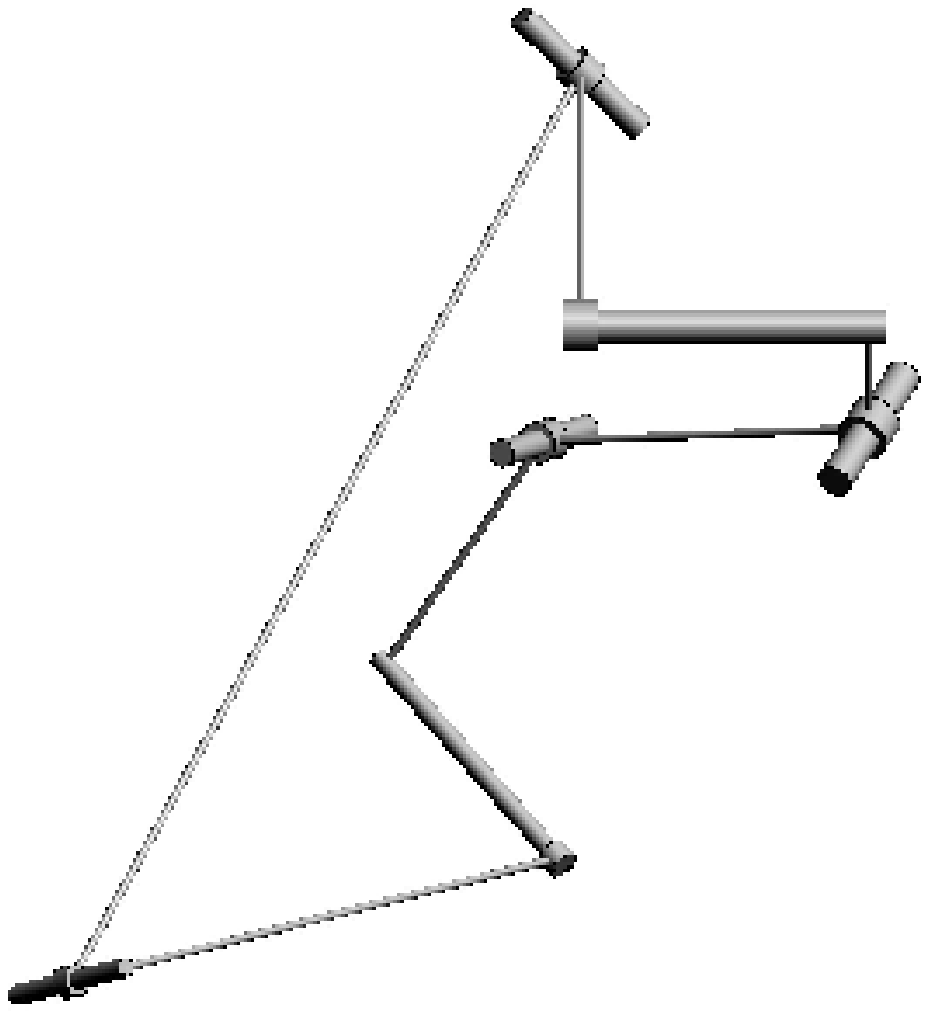}
                \caption{}
                \label{fig:b}
        \end{subfigure}
        \begin{subfigure}[b]{0.250\textwidth}
                \centering
                \includegraphics[width=\textwidth]{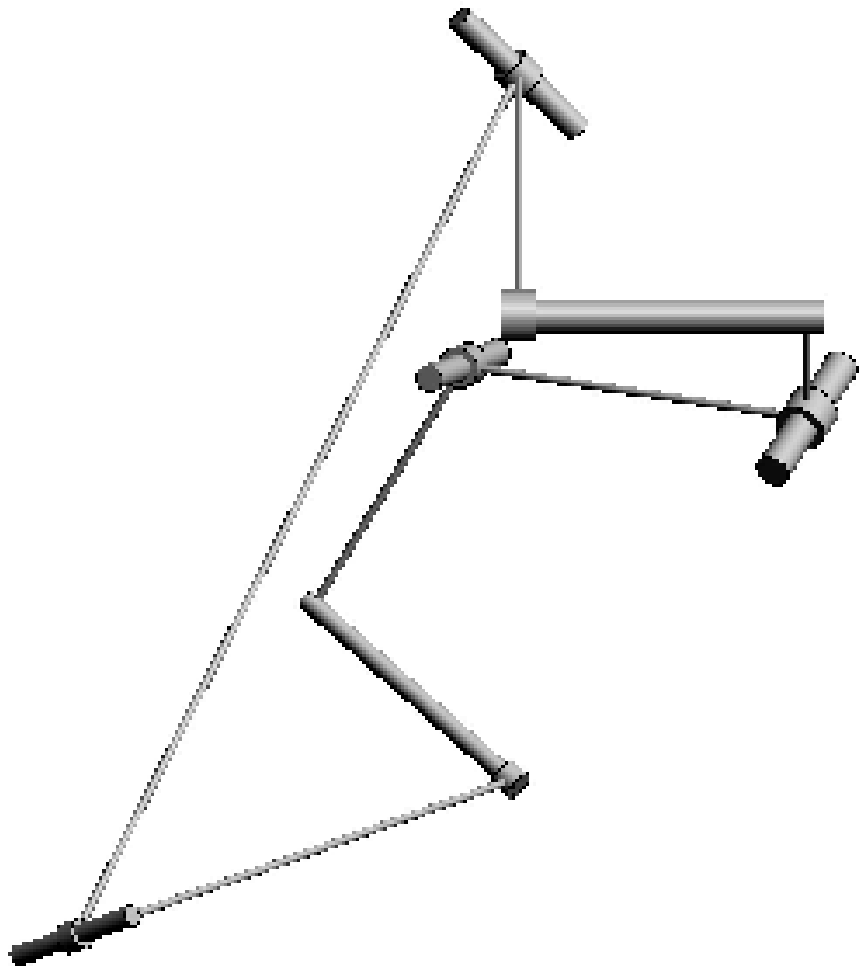}
                \caption{}
                \label{fig:c}
        \end{subfigure}
        \caption{Three configurations of the new linkage.
	}\label{fig:m}
\end{figure*}

\begin{thm}
The solution above is a set of Denavit-Hartenberg-parameters for a mobile linkage.
This linkage is different from all linkages listed in \cite{chenr12}.
\end{thm}

\begin{pf}
We calculate the configuration space using the method described in Section~\ref{pr:0}.
The Gr\"obner basis consists of 32 polynomials (1194 terms in total) in $t_1,\dots,t_6$ and the radical number $w_5$, 
 which are too long
to be reproduced here. The dimension of the ideal can be calculated from the Gr\"obner bases,
and it is indeed 1 and contains infinite real solutions. This shows that the linkage moves. Some comfigurations
are shown in Figure~\ref{fig:m}.

In order to show that the linkage is different from the known linkages in lists \cite{B93, Dietmaier95, chenr12}, 
one could compute all bond diagrams of the known linkages and see that the diagram in Figure~\ref{fig:bonds}(c)
is not among them. (Indeed, this was our first proof.) The disadvantage of this approach is that
we would have to include the bond diagrams of all known linkages. However, there is a shortcut based
on the observation that almost all linkages in \cite{B93, Dietmaier95, chenr12} fulfill at least one Bennett condition, while
our new Example does not satisfy Bennett conditions. We just need to check against the known
examples that do not fulfill the Bennett conditions. There are the Bricard line symmetric linkage,
the Bricard orthogonal linkage, and the cube linkage. For these four cases, the bond diagrams
are Figure~\ref{fig:bonds}(b),(h), and (e), and these are clearly different from Figure~\ref{fig:bonds}(c).
\end{pf}

One can observe, in addition, that all points in the configuration space satisfy the equation
$t_1=t_4$. When we vary $s_1=s_4$, we get a configuration set of a CRRCRR linkage (2 cylindrical
joints). This set is an infinite union of curves, so its dimension is 2. Let $a\in\R$.
Then we may pose a new constraint like $t_1=\cot(as_1)$ as \cite{ALS14}, and still have dimension 1; 
the second constraint $t_4=\cot(as_4)$ is implied because $t_1=t_4$ and $s_1=s_4$. 
This defines an HRRHRR linkage (2 helical joints) with movability~1.

\section{Acknowledgements} 
 We would like to thank G\'abor Heged\"us and Hans-Peter Schr\"ocker
  for discussion and helpful remarks.
 The research was supported by the Austrian Science Fund (FWF):
W1214-N15, project DK9.

\end{document}